\documentclass[conference]{IEEEtran}
\IEEEoverridecommandlockouts
\usepackage{cite}
\usepackage{amsmath,amssymb,amsfonts}
\usepackage{algorithmic}
\usepackage{graphicx}
\usepackage{textcomp}
\usepackage{xcolor}
\usepackage[hidelinks]{hyperref}
\usepackage[utf8]{inputenc}
\usepackage{booktabs}
\usepackage{multirow}
\usepackage{subfigure}
\def\BibTeX{{\rm B\kern-.05em{\sc i\kern-.025em b}\kern-.08em
    T\kern-.1667em\lower.7ex\hbox{E}\kern-.125emX}}
\begin{document}

\title{Visual and Semantic Prototypes-Jointly Guided CNN for Generalized Zero-shot Learning\\
\thanks{}
}

\author{\IEEEauthorblockN{Chuanxing Geng, Lue Tao, and Songcan Chen}
\IEEEauthorblockA{\textit{College of Computer Science \& Technology} \\
\textit{ Nanjing University of Aeronautics \& Astronautics}\\
Nanjing, China\\
\{gengchuanxing, tlmichael, s.cheng\}@nuaa.edu.cn}
}

\maketitle

\begin{abstract}
  In the process of exploring the world, the curiosity constantly drives humans to cognize new things. \emph{Supposing you are a zoologist, for a presented animal image, you can recognize it immediately if you know its class. Otherwise, you would more likely attempt to cognize it by exploiting the side-information (e.g., semantic information, etc.) you have accumulated.} Inspired by this, this paper decomposes the generalized zero-shot learning (G-ZSL) task into an open set recognition (OSR) task and a zero-shot learning (ZSL) task, where OSR recognizes seen classes (if we have seen (or known) them) and rejects unseen classes (if we have never seen (or known) them before), while ZSL identifies the unseen classes rejected by the former. Simultaneously, without violating OSR's assumptions (only known class knowledge is available in training), we also first attempt to explore a new generalized open set recognition (G-OSR) by introducing the accumulated side-information from known classes to OSR. For G-ZSL, such a decomposition effectively solves the class overfitting problem with easily misclassifying unseen classes as seen classes. The problem is ubiquitous in most existing G-ZSL methods. On the other hand, for G-OSR, introducing such semantic information of known classes not only improves the recognition performance but also endows OSR with the cognitive ability of unknown classes. Specifically, a visual and semantic prototypes-jointly guided convolutional neural network (VSG-CNN) is proposed to fulfill these two tasks (G-ZSL and G-OSR) in a unified end-to-end learning framework. Extensive experiments on benchmark datasets demonstrate the advantages of our learning framework.
\end{abstract}


\section{Introduction}
Under a closed set of classes (or static environment) assumption, the traditional recognition/classification algorithms have already achieved significant success in a variety of machine learning tasks. However, the more realistic scenario is usually open and non-stationary, where unseen (or unknown\footnote{The concepts of 'unseen class' and 'unknown class' are respectively from G-ZSL and OSR tasks. The 'unseen class' in G-ZSL denotes the class with no available instances in training, but available semantic information about it, while 'unknown class' in OSR represents the class without any information regarding it, i.e., there are neither training instances nor other side-information about it. }) classes can emerge unexpectedly. To meet this challenge, generalized zero-shot learning (G-ZSL) \cite{socher2013zero,chao2016empirical} and open set recognition (OSR) \cite{scheirer2013toward,scheirer2014probability,yang2018robust,geng2018recent,dhamija2018reducing,oza2019c2ae} recently have been widely explored. In G-ZSL, only the instances of seen classes and the semantic information (including seen and unseen classes) are informed during training. The learned classifiers need to recognize both seen and unseen classes, where they leverage semantic information sharing to bridge the seen and unseen classes. Compared to G-ZSL, OSR probably faces more serious challenge due to the fact that only the information from known classes is available and nothing about unknown classes. Similar to G-ZSL, the learned classifiers need to not only accurately classify the known classes but also effectively deal with the unknown ones. In this paper, we mainly focus on these two hot issues at the moment.

\begin{figure*}[!t]
\centering
\subfigure[Representation in \textbf{Semantic Space}]{\includegraphics[width=0.45\textwidth]{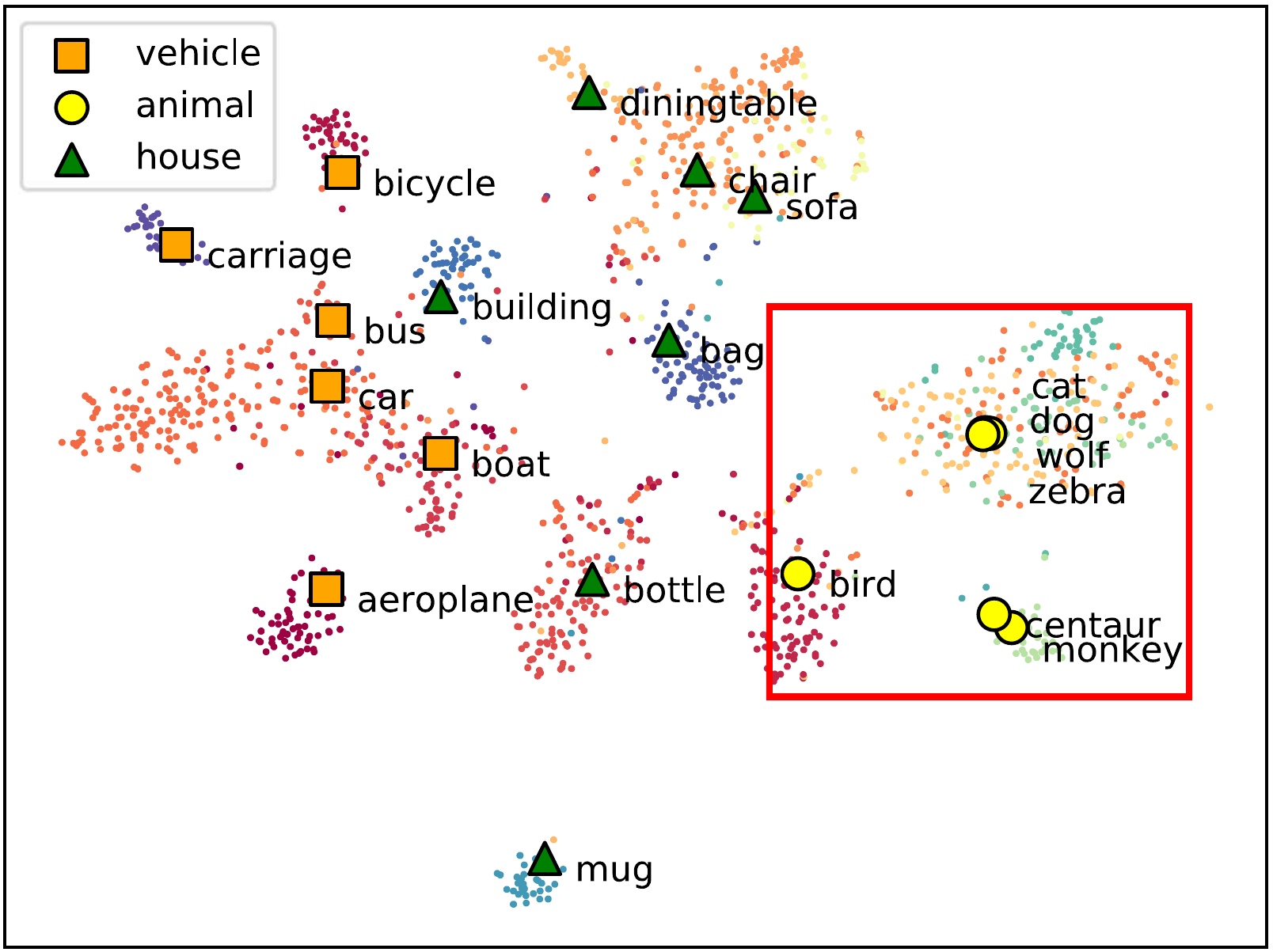}%
\label{fig_first_case}}
\hfil
\subfigure[Representation in \textbf{Visual Space}]{\includegraphics[width=0.45\textwidth]{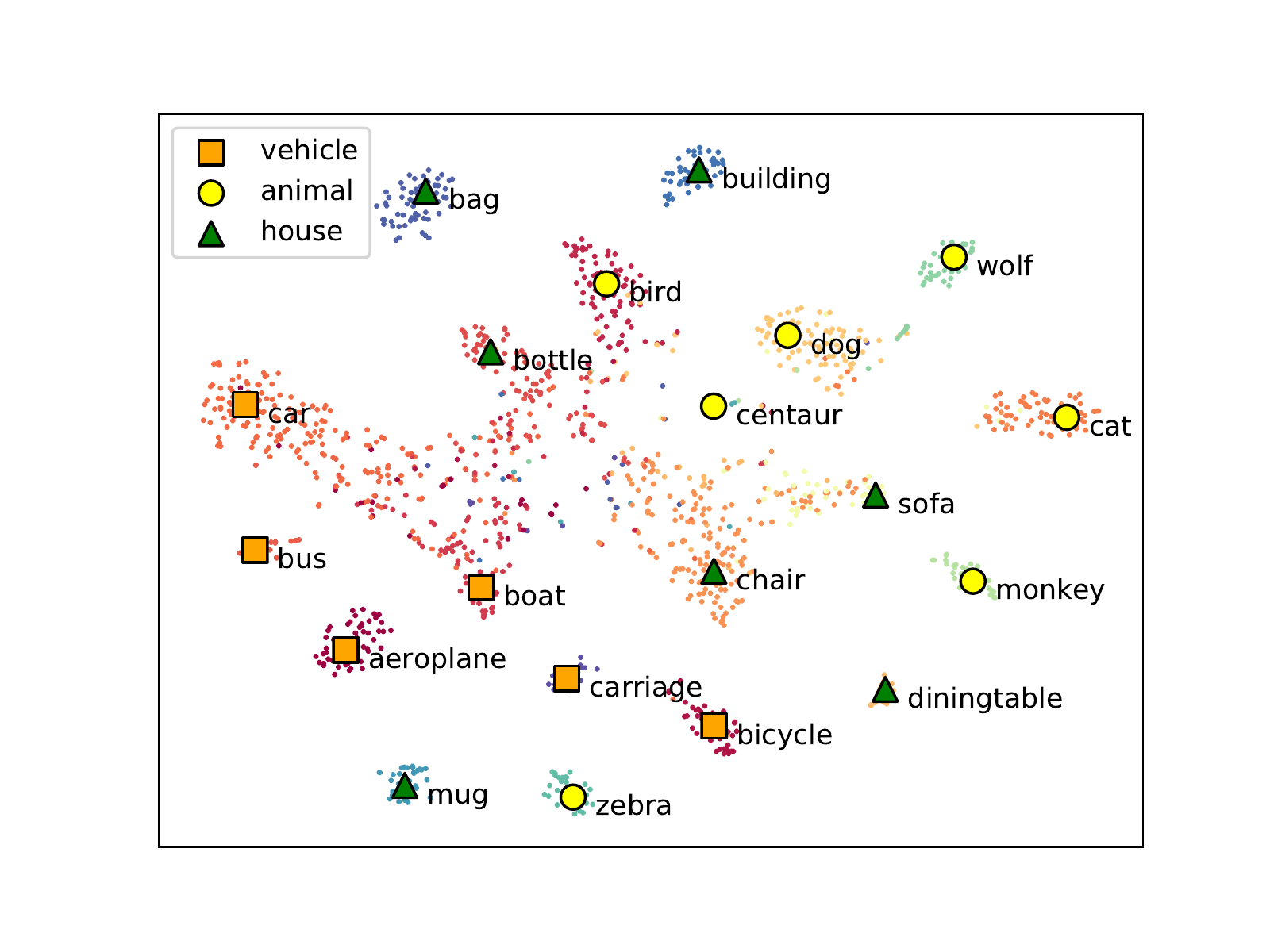}%
\label{fig_second_case}}
\caption{Visualization of the seen-class prototypes on validation set of aPY dataset in semantic space (a) and visual space (b) by t-SNE, where the prototypes in semantic space (semantic prototypes) are predefined, while the prototypes in visual space (visual prototypes) are learned by a convolutional prototype network \cite{yang2018robust}. To make it intuitive, the classes are further divided into three groups, i.e., vehicle, animal and house.}
\label{fig_sim}
\end{figure*}

\textbf{For G-ZSL}, one of the most serious challenges is the class overfitting (CO) problem \cite{zhang2019triple}, i.e., the seen class overfitting problem where the learned classifiers easily misclassify unseen classes as seen classes due to unavailable unseen class instances in training. Most existing methods usually mix seen and unseen classes together for recognition. They either learn a suitable visual-semantic embedding space \cite{rahman2018unified,liu2018zero,zhu2019generalized,zhang2019triple,chao2016empirical,liu2018generalized} or transform this problem to a traditional supervised problem by generating the synthetic instances for unseen classes using the variational autoencoder (VAE) or the generative adversarial networks (GAN) \cite{schonfeld2018generalized,xian2018feature,felix2018multi}.

For the visual-semantic embedding methods, although learning the space is crucial for G-ZSL \cite{Zhang2018CVPR}, mixing the seen and unseen classes together for recognition makes these methods unable to effectively solve the CO problem, even if introducing some calibration terms to balance these two conflicting forces. Furthermore, just as the saying goes: A picture (visual information) is worth a thousand words (semantic information), recognizing seen classes in such an embedding space may also lose partial discriminability of seen classes since the representation in visual space is usually richer and more discriminative than the counterpart in semantic space. As shown in Fig. 1, although the semantic representation indeed has good discriminability at 'vehicle-animal-house' class level, the discriminability of the subclasses in animal group suffers a serious decline. In contrast, the visual representation performs well at both 'vehicle-animal-house' class level and subclass-level. Similarly, for the methods of generating synthetic instances, although these methods greatly weaken the impact of the CO problem, this problem still exists due to the unreliability of these synthetic instances. These methods actually are limited by the generative techniques themselves, such as the blurriness of generative instances in VAE, the mode collapse and unstable training in GAN, etc \cite{goodfellow2016nips}.

\textbf{For OSR}, almost all existing methods focus on learning more robust classifiers to meet the challenge from the unknown classes during testing, where they adopt a reject operation to address unknown class instances. These methods currently just use the information of known classes at feature level, leaving out other side-information (such as semantic information, etc.) which is usually co-occurred with the visual feature information. As a result, the factors above limit the OSR!

Recall that the process of human exploring the world, the curiosity constantly drives us to cognize new things. Suppose that you are a zoologist and have an animal picture on your hand, you can recognize it immediately if you are familiar with this kind of animals. Otherwise, you want more likely to cognize or describe it by exploiting the side-information(e.g., semantic/attribute information) you have accumulated. In other words, you would not first figure out the abstract attribute of the animal and then classify it, if you are already familiar with this kind of animal.  However, that is exactly what the current visual-semantic embedding methods do: figure out the semantic vectors of instances at first, then classify them according to the nearest neighbor rule.

Inspired by this, we decompose G-ZSL into an OSR and a ZSL tasks, where OSR recognizes seen classes (if we have seen (or known) them) and rejects unseen classes (if we have never seen (or known) them before), while ZSL identifies the unseen classes rejected by the former. On the other hand, without violating OSR's assumptions, we also first attempt to explore a new generalized open set recognition (G-OSR) task by introducing the accumulated side-information from known classes. For G-ZSL, such a decomposition, i.e., identifying seen and unseen classes separately, effectively solves the CO problem. In addition, this operation also makes the recognition of seen classes in visual feature space, probably avoiding the loss of partial discriminability of seen classes as mentioned above. For G-OSR, introducing such semantic information of known classes not only improves the recognition performance but also endows OSR with the cognitive ability of unknown classes.

Specifically, a visual and semantic prototypes-jointly guided convolutional neural network (VSG-CNN) is proposed to fulfill these two tasks (G-ZSL and G-OSR) in a unified end-to-end learning framework. Note that although the G-ZSL is decomposed into an OSR and a ZSL tasks, our VSG-CNN can still make these two tasks jointly learned in an end-to-end framework, consequently, leading to improved experimental results. Furthermore, such a jointly learning mechanism intuitively also makes the visual and semantic information complement each other and benefit together. Fig. 2 shows the learning framework of our VSG-CNN. Our main contributions can be highlighted as follows:
\begin{enumerate}
\item Decomposing G-ZSL task into an OSR task and a ZSL task effectively addresses the CO problem which is ubiquitous in most existing G-ZSL methods.
\item Introducing the accumulated side-information from known classes to OSR to first explore a new generalized open set recognition (G-OSR) task.
\item A visual and semantic prototypes-jointly guided convolutional neural network (VSG-CNN) is proposed to fulfill these two tasks (G-ZSL and G-OSR) in a unified end-to-end learning framework.
\item Extensive experiments on benchmark datasets indicate the validity of our VSG-CNN.
\end{enumerate}

The rest of this paper is organized as follows. In Section II, we simply review the related works. Section III introduces the details of our learning framework. Extensive experimental results and analysis are reported in Section IV. Section V concludes this paper.

\begin{figure*}[!t]
  \centering
  \includegraphics[width=0.7\textwidth]{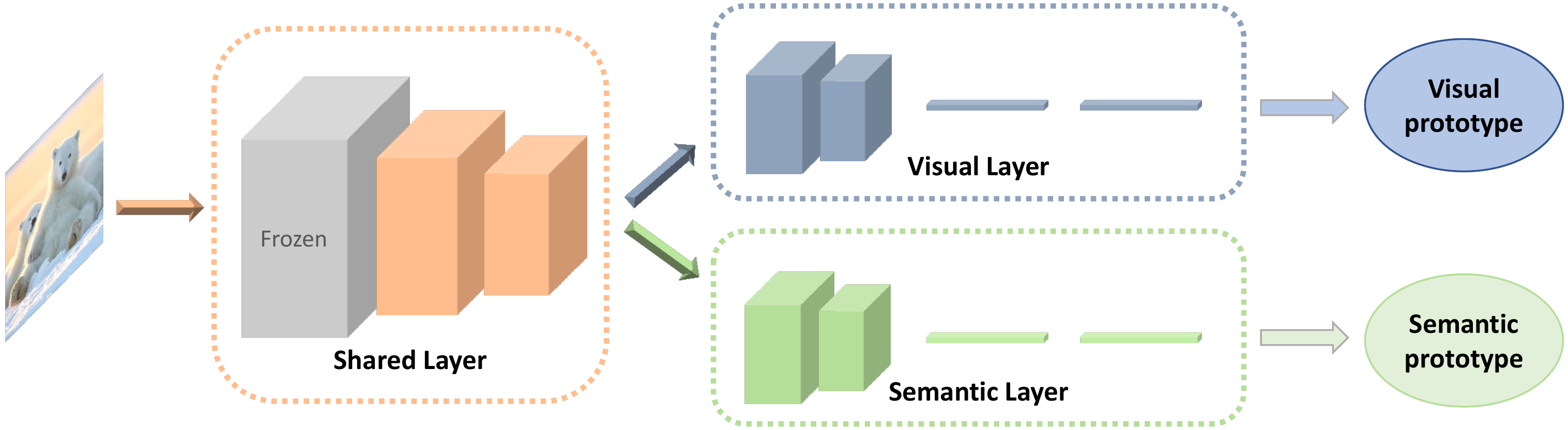}
  \caption{Overview of the VSG-CNN framework. The Shared Layer and the Visual Layer constitute the convolutional prototype subnetwork as one branch, where the visual prototype is learnable, while the Shared Layer and the Semantic Layer form the simple visual-semantic embedding subnetwork as other branch, where the semantic/attribute prototype is predefined. To prevent overfitting, the part of the Shared Layer is frozen during training.}
  \label{fig_sim}
\end{figure*}

\section{Related Work}
\subsection{(Generalized) Zero-shot Learning}
Inspired by the ability of humans to recognize without seeing examples, zero-shot learning (ZSL) \cite{lampert2009learning,lampert2013attribute,li2015semi,xian2017zero,fu2018recent,zhao2018zero} has been extensively studied, which leverages semantic information to bridge the seen and unseen classes. Existing ZSL task mainly focuses on the unseen classes' recognition, i.e., it often assumes that the testing instances only come from unseen classes. However, such an assumption is rather restrictive and impractical, since we usually know nothing about the testing instances either from seen classes or unseen classes. In addition, the object frequencies in natural images ordinarily follow long-tailed distributions \cite{salakhutdinov2011learning,zhu2014capturing}, meaning that the seen classes are more common than the unseen ones.

Expanded from ZSL, generalized zero-shot learning (G-ZSL) is a more realistic and challenging task, which needs to effectively predict both the seen and unseen classes, but has the same training settings in ZSL. Recently proposed G-ZSL methods can be roughly divided into three categories: \emph{visual-semantic embedding} (VS-Embedding), \emph{visual data augmentation} (VD-Augmentation), and \emph{domain separating} (DS).

{\bf Visual-Semantic Embedding}.
Since the key to the recognition of unseen class is the sharing of semantic information between them and seen classes, many G-ZSL methods mainly focus on learning a more suitable visual-semantic embedding space according to the mapping from seen class visual features to their corresponding semantic information/prototypes \cite{zhang2015zero,romera2015embarrassingly,xian2016latent,kodirov2017semantic,zhang2017learning,chao2016empirical,Zhang2018CVPR,guo2018implicit,chen2018zero,liu2018zero,zhang2018towards,annadani2018preserving,zhu2019generalized,zhang2019dual,yang2018dissimilarity,rahman2018unified,liu2018generalized,zhang2019triple}.
When a testing instance (either from a seen or an unseen class) arrives, it is first mapped into such a space, then classified according to the nearest neighbor rule. Note that due to no available unseen class instances, such learned classifiers are susceptible to incur CO problem. Although some calibration terms have been introduced \cite{chao2016empirical,liu2018generalized} to balance these two conflicting forces, the CO problem actually has not been effectively solved.

{\bf Visual Data Augmentation}.
To overcome the absence of unseen class instances in training, the VD-Augmentation methods \cite{xian2018feature, long2018pseudo,verma2018generalized,felix2018multi,yu2018bi,li2019leveraging,schonfeld2018generalized,felix2019multi} mainly rely on the generative models (such as VAE or GAN) to generate their synthetic instances according to the corresponding semantic prototypes. Thus the classifiers can be trained based on both seen class instances and unseen class synthetic instances, turning G-ZSL into a traditional supervised learning task. Such methods currently have achieved remarkable experimental results. However, the synthetic instances for unseen classes are not always reliable due to the limitations of the generative technique, such as the blurriness of generative instances in VAE, the mode collapse and unstable training in GAN, etc. Therefore, the CO problem still exists.



{\bf Domain Separating}.
Treating the seen and unseen classes separately is the main idea of DS methods  \cite{socher2013zero,dong2018learning,atzmon2019adaptive}, where the seen classes were classified by a traditional classifier while a ZSL classifier for unseen classes. Thus, a detector used for distinguishing between seen and unseen classes is crucial. Intuitively, if the detector can accurately separate seen classes from unseen classes, this kind of methods will essentially solve the CO problem. Please note that most this kind of methods currently train the learners mentioned above \textbf{completely separating}, and such an operation is easy to incur a suboptimal solution. In contrast, though our decomposition on G-ZSL is similar in spirit to the DS idea, unlike the separately training process in DS, our VSG-CNN is an \textbf{end-to-end jointly learning} framework, where the visual and semantic prototypes jointly guide the CNN learning. Such a learning manner intuitively can also make the visual and semantic information complementary and benefit from each other.

%
%


\subsection{Open Set Recognition}
Open set recognition (OSR)\cite{scheirer2013toward,scheirer2014probability,yang2018robust,geng2018recent,dhamija2018reducing,oza2019c2ae} describes such a scenario where new classes unknown in training appear in testing, requiring the classifiers to not only accurately classify the known/seen classes but also effectively address the unknown ones. Existing OSR methods place more emphasis on identifying known/seen classes, especially when unknown classes appear in testing. This makes them have a reject operation for unknown classes. Almost all existing OSR methods currently just use the information of seen classes at visual feature level, whereas their corresponding side-information (such as semantic information, etc.) is completely left out. With the exploration of ZSL, we can find that a lot of semantic information is usually shared between the known and unknown classes. Intuitively, exploiting this information  not only assists the seen classes' recognition but also serves to cognize unknown classes. Therefore, this paper also tries to explore the effective exploitation of semantic information in OSR.

\section{Proposed Approach}
As discussed in Section I, this paper mainly focuses on G-ZSL and OSR, which are the hot issues at the moment. Inspired by the process of humans exploring the world, we strategically combine the OSR and ZSL methods together to solve the G-ZSL task. This effectively solves the CO problem. On the other hand, we also first attempt to explore a new generalized open set recognition (G-OSR) by introducing the accumulated semantic information from known classes to OSR. Specifically, a visual and semantic prototypes-jointly guided convolutional neural network (VSG-CNN) is proposed to fulfill these two tasks (G-ZSL and G-OSR) in a unified end-to-end learning framework. Next, we will provide more details in the following subsections.


\subsection{Problem Formulation}
Let $\mathcal{D}_s=\{x_i,y_i,a_{y_i}\}_{i=1}^N$ denote the training set which has $N$ instances from seen classes, where $x_i\in \mathbb{R}^d$ is the visual feature of $i$-th instance with the class label $y_i\in \mathcal{C}_s$ ($\mathcal{C}_s$ represents the seen class set). Let $A_s=\{a_{y_i}|y_i\in \mathcal{C}_s\}$ represent the semantic/attribute set of seen classes. In general, the $a_{y_i}$ of instances in one class should be the same. Similarly, let $\mathcal{C}_u$ and $A_u=\{a_{y_i}|y_i\in \mathcal{C}_u\}$ respectively denote the class and attribute sets of unseen classes, where $\mathcal{C}_s\cap \mathcal{C}_u=\emptyset$. Given one testing instance $x$, the goal of classification is to predict its class label $y$. Thus we have:
\begin{enumerate}
\item[1)] for the ZSL task, $y\in \mathcal{C}_u$;
\item[2)] for the G-ZSL task, $y\in\mathcal{C}_s\cup \mathcal{C}_u$;
\item[3)] for the OSR task, $y\in\{\mathcal{C}_s,\text{'unknown class'}\}$;
\item[4)] for the new G-OSR task, $y\in\{\mathcal{C}_s,\text{'unknown class'}\}$ with cognizing unknown classes using the semantic/attribute from known/seen classes.
\end{enumerate}


\subsection{Convolutional Prototype Learning}
Due to the fact that the softmax layer in convolutional neural networks (CNNs) is based on the assumption of closed world (i.e., with a fixed number of classes), the CNNs usually lack robustness in the real-world classification/recognition applications, where incomplete knowledge of the world often exists during training and while unknown classes can be submitted to a learned algorithm during testing. To strengthen the robustness, \cite{yang2018robust} proposed a convolutional prototype network (CPL) where it replaces the traditional cross-entropy loss with distance based cross entropy loss (DCE), or margin based classification loss (MCL), or minimum classification error loss (MCE) and prototype loss (PL). In fact, CPL learns several prototypes on the visual features (here we call them \emph{visual prototypes}) for each class, making it handle the OSR task well. Note that similar to semantic/attribute prototype, the \emph{visual prototypes} of instances in one class should be the same, while for convenience we here represent each class with one visual prototype. Furthermore, we adopt the DCE and PL losses in this paper.

In the CPL framework, the probability of an instance $(x_i, y_i)$ belonging to its \emph{visual prototype} $m_{y_i}\in \mathbb{R}^t$ can be defined as
\begin{equation}
p(x_i\in m_{y_i}|x_i) = \frac{e^{-\gamma d(f(x_i),m_{y_i})}}{\sum_{k=1}^Ce^{-\gamma d(f(x_i),m_{k})}},
\end{equation}
where $f(x_i)$ is the CNN feature extractor, $d(f(x_i),m_{y_i})= \|f(x_i)-m_{y_i}\|_2^2$ represents the distance between $f(x_i)$ and $m_{y_i}$, and $C$ and $\gamma$ respectively denote the number of training classes and a hyper-parameter that controls the hardness of probability assignment. Let $Z=f(x_i)$ and $M=m_{y_i}$, thus the DCE loss can be defined as:
\begin{equation}
L_{\text{DCE}}(Z,M) = -\log p(x_i\in m_{y_i}|x_i).
\end{equation}

To further improve the generalization performance of CPL, the authors again introduced the PL loss $L_{\text{PL}}(Z,M) = \|f(x_i)-m_{y_i}\|_2^2$ as a regularization term. Then the total loss of CPL is as follows:
\begin{equation}
L(Z,M) = L_{\text{DCE}}(Z,M) + \lambda L_{PL}(Z,M),
\end{equation}
where $\lambda$ is a regularization parameter.

\subsection{VSG-CNN Learning Framework}
\subsubsection{VSG-CNN}
Our proposed VSG-CNN learning framework decomposes the G-ZSL task into an OSR and a ZSL tasks, and its core idea is making these two decomposed tasks able to be jointly trained to benefit each other. To corroborate the validity of such a decomposition, we simply append a semantic convolutional prototype subnetwork (SCPN, implementing the ZSL task) to CPL to constitute our VSG-CNN. Note that SCPN can be replaced with other complicated ZSL methods, but beyond our focus here. Thus, we adopt both visual and semantic prototypes to jointly guide the convolutional neural network learning. As shown in Fig. 2, the Shared Layer and the Visual Layer constitute the convolutional prototype subnetwork used for OSR, where the visual prototype is learnable, while the Shared Layer and the Semantic Layer form the visual-semantic embedding subnetwork used for ZSL, where the semantic/attribute prototype is predefined. To prevent overfitting, the part of the Shared Layer is frozen during training.

Specifically, let $Z_{\text{shared}} = f_{\text{shared}}(x_i)$ denote the shared visual feature from the \emph{Shared Layer}, $Z_{v} = f_v(Z_{\text{shared}})$ and $Z_{s}=f_s(Z_{\text{shared}})$ respectively represent the features extracted from the \emph{Visual Layer} and the \emph{Semantic Layer}, and $M_v$ denotes the visual prototype, the whole optimization objective of VSG-CNN is defined as follows:
\begin{equation}
\min L(Z_v,M_v) + L(Z_s,A_s),
\end{equation}
where the first term indicates the visual prototype learning loss similar to (3), and the second term is the visual-semantic mapping learning loss. At first glance, the formula (4) seems to have the same form as (3) in CPL \cite{yang2018robust}. However, please note that it is essentially different from CPL: (i) we jointly use visual and semantic prototypes to guide the network's learning; (ii) we predefine the semantic prototype, making it not need to be learned.

\subsubsection{Prediction}
\begin{figure*}[!t]
\centering
\subfigure[]{\includegraphics[width=0.45\textwidth]{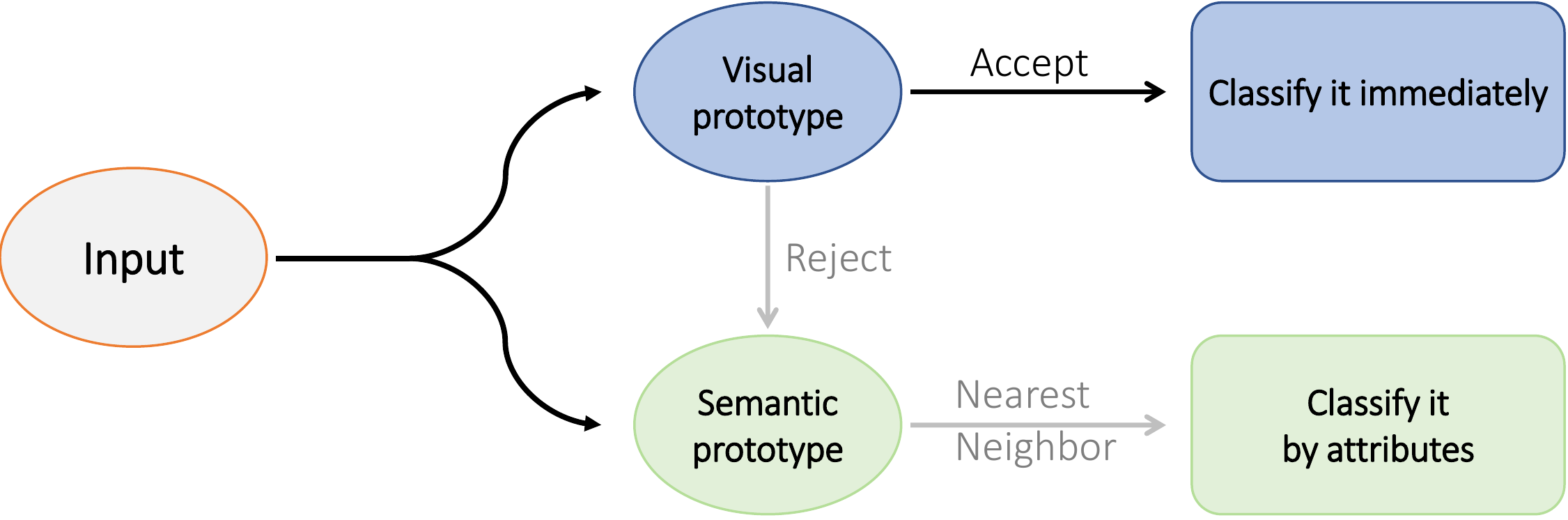}%
\label{fig_first_case}}
\hfil
\subfigure[]{\includegraphics[width=0.45\textwidth]{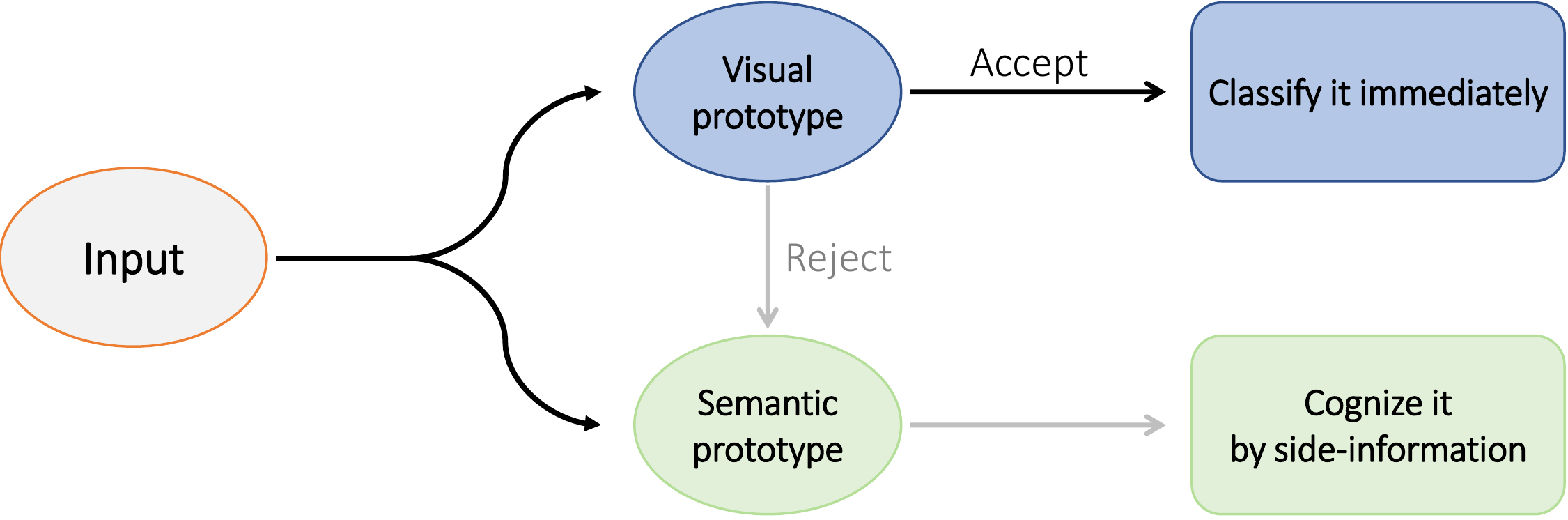}%
\label{fig_second_case}}
\caption{(a) and (b) respectively describe the prediction processes in G-ZSL and G-OSR. Different from G-ZSL, nothing about the unknown classes is available in G-OSR. Therefore, we humans only rely on the knowledge/side-information gained from known classes to cognize unknown classes.  }
\label{fig_sim}
\end{figure*}
This paper involves two learning tasks, i.e., the G-ZSL task and the G-OSR task. \textbf{For the G-ZSL task}, when a testing instance arrives, VSG-CNN will first achieve its representations of both the visual and semantic prototypes. Then the formula (1) is used to obtain the probability score set of this instance corresponding to each seen class visual prototype, i.e., $\{p(x\in m_{y_s}|x)|y_s\in{\mathcal{C}_s}\}$, and these scores will in turn be used for determining which domain (i.e., seen class or unseen class) this instance belongs to by calculating their entropy value $E=-\sum p(x\in m_{y_s}|x)\log p(x\in m_{y_s}|x)$: if $E$ is less than a predefined threshold $\delta_{g}$, it will be labeled as some seen class closest to its visual prototype representation according to the nearest neighbor rule, otherwise, as some unseen class in the same way based on its semantic/attribute prototype representation. Fig. 3(a) describes this prediction process. \textbf{Note that} such operations make predictions of seen and unseen classes separately, effectively handling the CO problem. Meanwhile, identifying seen classes in visual space may lead to better recognition performance as discussed in Section I.


\textbf{For the G-OSR task}, similar to the predicting process of G-ZSL, we first obtain the entropy value of the corresponding testing instance. Then, if the value is less than a predefined threshold $\delta_o$, it will be labeled as some seen class closest to its visual prototype representation, otherwise as an unknown class, at the same time also output its semantic/attribute prototype representation. Fig. 3(b) describes such a prediction process. \textbf{Note that} the main differences between VSG-CNN and other existing OSR methods are: (i) when identifying known classes, VSG-CNN first attempts to use the semantic/attribute information (from known classes) obviously ignored by other existing OSR methods; (ii) instead of simply rejecting unknown classes, VSG-CNN can also output its corresponding semantic prototype representation, which is critical for humans to cognize new things. This details in Section IV.

\section{Experiments and Analysis}
\subsection{Datasets and Training Details}
\subsubsection{Datasets}
Following the data splits proposed by \cite{xian2018zero}, we assess our VSG-CNN on four benchmark datasets  as follows:\\
\textbf{CUB}\cite{wah2011caltech}: i.e., Caltech-UCSD Birds-200-2011, has a total of 11,788 fine-grained images from 200 bird species. Each species is annotated with 312 attributes. It contains 150 seen (50 validation classes) and 50 unseen classes.\\
\textbf{AWA2}\cite{xian2018zero}: has a total of 37,322 coarse-grained images from 50 animals. Each animal is annotated with 85 attributes. It has 40 seen (13 validation classes) and 10 unseen classes.\\
\textbf{SUN}\cite{patterson2012sun}: has a total of 14,340 fine-grained scene images from 717 scene classes. Each scene class is annotated with 102 attributes. It contains 645 seen (65 validation classes) and 72 unseen classes.\\
\textbf{aPY}\cite{farhadi2009describing}: i.e., aPascal \& aYahoo, has a total of 12,051 coarse-grained images from 32 generic object classes. Each class is annotated with 64 attributes. It has 20 seen (5 validation classes) and 12 unseen classes.\\
Additionally, Table I describes these datasets.

\begin{table}
\centering
\caption{Statistics of the four benchmark datasets.}
\begin{tabular}{lcccc}
\toprule
\textbf{Dataset}& \# Attr &\# Seen &\# Unseen&\# Imag \\
\midrule
 \textbf{CUB}    & 312     &100 + 50& 50      & 11788 \\
 \textbf{AWA2}    & 85      &27 + 13 & 10      & 37322 \\
 \textbf{SUN}    & 102     &580 + 65& 72      & 14340 \\
 \textbf{aPY}    & 64      &15 + 5  & 12      & 15339 \\
\bottomrule
\end{tabular}

\label{tab:booktabs}
\end{table}


\begin{figure}[!t]
  \centering
  \includegraphics[width=0.4\textwidth]{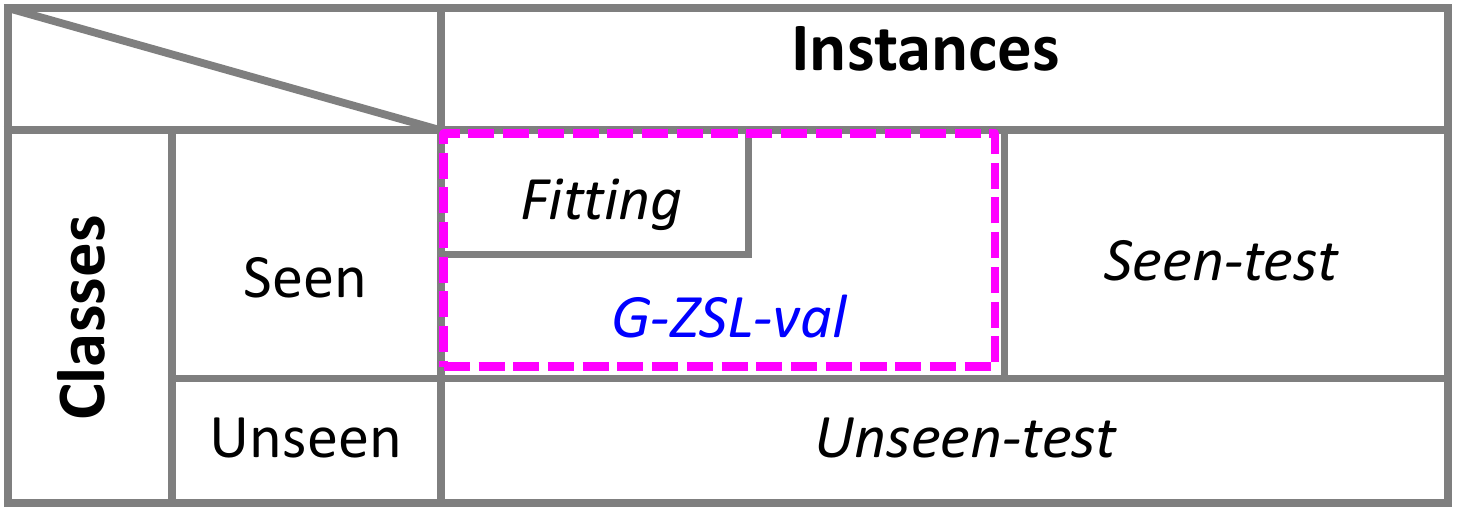}
  \caption{G-ZSL-val split. The data is organized across classes and instances. We randomly select 6/7 of seen-training set as Fitting set, while the remaining 1/7 of seen-training set and the seen-validation set constitute the G-ZSL-val set (the seen-training and seen-validation sets are proposed by \cite{xian2018zero}.)}
  \label{fig_sim}
\end{figure}

\begin{table*}[t]
\centering
\caption{Comparing CRL with state-of-art G-ZSL visual-semantic embedding and domain separating methods. '-' indicates the corresponding methods do not provide their results or the datasets they do not experiment. Best $H$ results (\%) are indicated in bold.}
\tabcolsep 2.4mm
\renewcommand\arraystretch{1.27}
\begin{tabular}{l|ccc|ccc|ccc|ccc}
\hline
\multirow{2}{*}{Method / Dataset} & \multicolumn{3}{c|}{\textbf{CUB}} & \multicolumn{3}{c|}{\textbf{AWA2}} & \multicolumn{3}{c|}{\textbf{SUN}} & \multicolumn{3}{c}{\textbf{aPY}} \\
                  &ts (\%)   & tr(\%) &   H (\%)  &  ts (\%) & tr (\%)  &  H (\%)    & ts (\%)   &   tr (\%) & H (\%) &   ts (\%)   &  tr (\%)    &   H (\%)  \\
                  \hline
  \textbf{Visual-Semantic Embedding}     &       &       &       &       &       &       &       &       &       &       &       &      \\
       SSE \cite{zhang2015zero}       &  8.5  & 46.9  & 14.4 & 8.1 &  82.6 &  14.8  &  2.1 & 36.4  &  4.0 & 0.2  &  78.9 & 0.4 \\
       ESZSL \cite{romera2015embarrassingly} &  12.6 &63.8 &21.0 & 5.9 &77.8 &11.0  & 11.0 & 27.9 & 15.8 & 2.4 &70.1 &4.6 \\
       LATEM \cite{xian2016latent}       &  15.2 & 57.3 & 24.0 & 11.5 & 77.3 & 20.0  &  14.7 & 28.8 & 19.5 & 0.1 & 73.0 & 0.2 \\
       SAE \cite{kodirov2017semantic}    &  7.8  &54.0  &13.6  & 1.1  &82.2  &2.2  & 8.8 &18.0 &11.8 & 0.4 &80.9 &0.9 \\
       DEM  \cite{zhang2017learning}     &  19.6 &57.9  &29.2 & 30.5 &86.4 &45.1  & 34.3 &20.5 &25.6   & 11.1 &79.4 &19.4  \\
       KERNEL \cite{Zhang2018CVPR}       &  19.9 &  52.5 &  28.9 & 17.6  & 80.9  & 29.0  &  19.8 & 29.1  &  23.6 & 11.9  &  76.3 & 20.5 \\
       ICINESS \cite{guo2018implicit}    &   -   &  -    &  41.8 & -     & -   & 41.0  &  -   & -   &  32.1 & -   & -     & 25.4 \\
       EDE  \cite{zhang2018towards}      &  21.0 &  66.0 &  31.9 & 35.2  & 93.0  & 51.1  &  22.1 & 35.6  &  27.3 & 7.8   &  75.3 & 14.1 \\
       DCN \cite{liu2018generalized}     &  28.4 &  60.7 &  38.7 & - & -  & -  & 25.5  & 37.0  & 30.2  & 14.2  &  75.0 & 23.9 \\
       PSR \cite{annadani2018preserving} &  24.6 &  54.3 &  33.9 & 20.7  & 73.8  & 32.3  &  20.8 & 37.2  &  26.7 & 13.5  &  51.4 & 21.4 \\
       VSE  \cite{zhu2019generalized}    &  33.4 &  87.5 &  48.4 & 41.6  & 91.3  & 57.2  &  -    &   -   &  -    & 24.5  &  72.0 & \textbf{36.6} \\
    $\text{DVN}$ \cite{zhang2019dual}  &  29.0 &  58.6 &  38.8 & -  & -  & -  & 20.8  & 31.0  & 24.9  & 24.5  &  56.1 & 34.1 \\
       DSS \cite{yang2018dissimilarity}  &  25.0 &  93.2 &  39.4 & 15.7  & 97.5  & 27.0  & 18.2  & 81.4  & 29.7  & 12.8  &  93.6 & 22.5 \\
       CAPDs \cite{rahman2018unified}    &  43.3 &  41.7 &  44.9 & -  & -  & -  &  27.8  & 35.8 & 31.3 & 37.0  &  59.5 & 26.8 \\
       RN  \cite{sung2018learning}       &  38.1 &  61.1 &  47.0 & 30.0  & 93.4  & 45.3  & -     &  -    &  -    & -     &   -   &  -   \\
       TRIPLE \cite{zhang2019triple}     &  26.5 &  62.3 &  37.2 & -  & -  & -  & 22.2  & 38.3  & 28.1  & 16.1  &  66.9 & 25.9 \\
       SP-AEN  \cite{chen2018zero}       &  34.7 &  70.6 &  46.6 & 23.3  & 90.9  & 37.1  &  24.9 & 38.6  &  30.3 & 13.7  &  63.4 & 22.6 \\
       LESAE \cite{liu2018zero}          &  24.3 &  53.0 &  33.3 & 21.8  & 70.6  & 33.3  &  21.9 & 34.7  &  26.9 & 12.7  &  56.1 & 20.1 \\
                  \hline
  \textbf{Domain Separating}&       &       &       &       &       &       &       &       &       &       &       &      \\
       CMT  \cite{socher2013zero}     & 7.2 &  49.8 & 12.6  & 0.5 & 90.0  & 1.0  & 8.1  & 21.8  & 11.8  &  1.4 & 85.2  & 2.8    \\
       CMT* \cite{socher2013zero}     & 4.7 &  60.1 & 8.7  & 8.7 & 89.0  & 15.9  & 8.7  & 28.0  & 13.3  &  10.9 & 74.2  & 19.0    \\
       COSMO  \cite{atzmon2019adaptive} &  44.4 & 57.8 & 50.2 & 54.9  & 76.2  & 63.8  & 44.9  & 37.7  & \textbf{41.0}  &  -    & -     & -    \\
       \hline
  Baseline (ours)                    &  46.5 &  64.5 &  54.1 & 52.0  & 74.7  & 61.3  & 24.8  & 34.1  & 28.7  & 22.2  & 64.7  & 33.0 \\
  VSG-CNN (ours)                         &  52.6 &  62.1 & \textbf{57.0} & 60.4  & 75.1  & \textbf{67.0} & 30.3  & 31.6  & 30.9  & 22.9  &  66.1 & 34.0 \\
                  \hline
\end{tabular}

\end{table*}

\subsubsection{Training Details}
\textbf{Architecture}: VSG-CNN\footnote{To encourage reproducible research, the code of our method will be
later published.} is based on ResNet-101 \cite{he2016deep} which takes a cropped $224\times224\times3$ image as input and outputs a 2048-dimensional visual feature vector. Specifically, the first 95 layers of our framework constitute the \emph{Shared Layer}, of which the first 92 layers are frozen to prevent the network overfitting. Moreover, the \emph{Visual Layer} contains the 96$\sim$101 layers of ResNet-101 and a fully-connected layer used for learning the visual prototypes, while the \emph{Semantic Layer} contains another 96$\sim$101 layers of ResNet-101 and a fully-connected layer used for mapping visual feature vectors to the corresponding semantic/attribute prototypes. Besides, since the visual prototypes are learnable, we randomly initialized them by a uniform distribution on the interval [0,1). In addition, we adopt the  image mirroring and image cropping data augmentation techniques commonly used in training a deep network. To further improve the efficiency, we also initialize our framework with the pre-trained ResNet-101 on ImageNet 1K \cite{russakovsky2015imagenet}.

\begin{table*}[t]
\centering
\caption{Comparing VSG-CNN with state-of-art G-ZSL visual data augmentation methods. '-' indicates the corresponding methods do not provide their results or the datasets they do not experiment. Best $H$ results (\%) are indicated in bold.}
\tabcolsep 2.4mm
\renewcommand\arraystretch{1.27}
\begin{tabular}{l|ccc|ccc|ccc|ccc}
\hline
\multirow{2}{*}{Method / Dataset} & \multicolumn{3}{c|}{\textbf{CUB}} & \multicolumn{3}{c|}{\textbf{AWA2}} & \multicolumn{3}{c|}{\textbf{SUN}} & \multicolumn{3}{c}{\textbf{aPY}} \\
                  &ts (\%)   & tr(\%) &   H (\%)  &  ts (\%) & tr (\%)  &  H (\%)    & ts (\%)   &   tr (\%) & H (\%) &   ts (\%)   &  tr (\%)    &   H (\%)  \\
                  \hline
  \textbf{Visual Data Augmentation}&       &       &       &       &       &       &       &       &       &       &       &      \\
       f-CLSWGAN  \cite{xian2018feature}        &  43.7 &  57.7 &  49.7 & 53.7 & 68.2 & 60.1 & 42.6  & 36.6  & 39.4  & 16.8  & 45.7  & 24.6    \\
       PTMCA  \cite{long2018pseudo}        &  23.0 &51.6 &31.8 & - & - & - & 19.0 &32.7 &24.0  & 15.4& 71.3 &25.4    \\
       SE-GZSL \cite{verma2018generalized}       &  41.5 &  53.3 &  46.7 & 58.3  & 68.1  & 62.8  & 40.9  & 30.5  & 34.9  & -     & -     & -    \\
     cycle-(U)WGAN \cite{felix2018multi}         &  47.9 &  59.3 &  53.0 & -  & -  & -  & 47.2  & 33.8  & 39.4  & -     & -     & -    \\
       BAAE   \cite{yu2018bi}                    &  -    &  -    &  -    & 51.4  & 85.6  & 64.2  & 23.1  & 36.7  & 28.4  & 15.4  &  74.1 & 25.5 \\
       LisGAN \cite{li2019leveraging}        &  46.5 & 57.9 & 51.6 & - & -  & -  & 42.9  & 37.8 & 40.2 & -  &  - & - \\
       CADA-VAE \cite{schonfeld2018generalized}  &  53.5 &  51.6 &  52.4 & 75.0  & 55.8  & 63.9  & 35.7  & 47.2  & \textbf{40.6}  & -     &  -    & -    \\
       3ME  \cite{felix2019multi}                &  49.6 &  60.1 &  54.3 & -  & -  & -  & 44.0  & 35.8  & 39.4  & -     &  -    & -    \\
                  \hline
  VSG-CNN (ours)                         &  52.6 &  62.1 & \textbf{57.0} & 60.4  & 75.1  & \textbf{67.0} & 30.3  & 31.6  & 30.9  & 22.9  &  66.1 & \textbf{34.0} \\
                  \hline
\end{tabular}

\end{table*}

\textbf{Parameter Selection}: Similar to \cite{atzmon2019adaptive}, we here introduce an additional split --- \emph{G-ZSL-val} shown in Fig. 4. We first train VSG-CNN on the Fitting set, then the grid search technique is adopted to select VSG-CNN's parameters  from the corresponding candidate sets on \emph{G-ZSL-val} set. Specifically, the thresholds $\delta_g$ and $\delta_o$ are selected from the interval of $[0:0.000002:0.02]$; $\lambda$ and the \emph{visual prototype} dimension $t$ are respectively selected from the candidate sets \{$10^{-3},10^{-2},10^{-1},10^{0},10^{1}$\} and \{$32,50,64,85,102,128,200,256,312,512,717$\}. Furthermore, the $\gamma$ in the convolutional prototype network of VSG-CNN is set to $1$. After determining these parameters, we retrain the VSG-CNN framework on the training set (in pink part of Fig. 4), and evaluate its performance on the four benchmark datasets.

\subsection{Results}
\subsubsection{Generalized Zero-shot Learning}
For the sake of fairness, we here focus on the comparisons of VSG-CNN with 20 leading G-ZSL methods which do not generate the synthetic instances of unseen classes. These methods respectively come from \emph{visual-semantic embedding} and \emph{domain separating}. Specifically, \textbf{VS-Embedding} methods contain SSE \cite{zhang2015zero}, ESZSL \cite{romera2015embarrassingly}, LATEM \cite{xian2016latent}, SAE \cite{kodirov2017semantic}, DEM \cite{zhang2017learning}, KERNEL \cite{Zhang2018CVPR}, ICINESS \cite{guo2018implicit}, SP-AEN \cite{chen2018zero}, LESAE \cite{liu2018zero}, EDE \cite{zhang2018towards}, PSR \cite{annadani2018preserving}, VSE \cite{zhu2019generalized}, DVN \cite{zhang2019dual}, DSS \cite{yang2018dissimilarity}, CAPDs \cite{rahman2018unified}, DCN \cite{liu2018generalized}, and TRIPLE \cite{zhang2019triple}. \textbf{Domain Separating} methods has the CMT \cite{socher2013zero} and COSMO \cite{atzmon2019adaptive}. Note that, we do not compare VSG-CNN with the method in \cite{dong2018learning} since its predictions used the information from testing set, and limited by the length of this paper, we refer the reader to \cite{dong2018learning} for more details. In addition, we also introduce a baseline method (Baseline) which adopts a non-co-learning strategy, i.e., the training of the convolutional prototype subnetwork and the visual-semantic embedding subnetwork of VSG-CNN is completely separate.

Besides, as a supplement, we also compare VSG-CNN and 8 leading \textbf{VD-Augmentation} G-ZSL methods including f-CLSWGAN \cite{xian2018feature},  PTMCA \cite{long2018pseudo}, SE-GZSL \cite{verma2018generalized}, cycle-(U)WGAN \cite{felix2018multi}, BAAE \cite{yu2018bi}, LisGAN \cite{li2019leveraging}, CADA-VAE \cite{schonfeld2018generalized}, 3ME \cite{felix2019multi}, which details in the following part. 


\textbf{Performance Evaluation}. Following \cite{xian2017zero}, we adopt the harmonic mean ($H$) of $tr$ --- the accuracy over seen classes, and $ts$ --- the accuracy over unseen classes as the evaluation metric, i.e, $$H = 2\times (ts\times tr)/(ts + tr).$$ Table 2 summaries the results of our VSG-CNN and the VS-Embedding and Domain Separating G-ZSL methods. Compared with VS-Embedding methods, the $H$ of our VSG-CNN achieves the significant improvements in CUB (\textbf{57.0}\% vs \textbf{48.4}\%) and AWA2 (\textbf{67.0}\% vs \textbf{57.2}\%), while it is comparable in SUN (\textbf{30.9}\% vs \textbf{32.1}\%) and aPY (\textbf{34.0}\% vs \textbf{36.6}\%). Furthermore, the gaps between
$ts$ and $tr$ in VSG-CNN on most datasets are smaller than VS-Embedding methods. This exactly indicates VSG-CNN's effectiveness, especially on CO problem. When compared with Domain Separating methods, although VSG-CNN fails in SUN (\textbf{30.9}\% vs \textbf{41.0}\%), it wins in CUB (\textbf{57.0}\% vs \textbf{50.2}\%), AWA2 (\textbf{67.0}\% versus \textbf{63.8}\%), and  aPY (\textbf{34.0}\% vs \textbf{19.0}\%). Additionally, our VSG-CNN is also consistently ahead of Baseline on all the four benchmark datasets, which indicates the effectiveness of the visual and semantic prototypes' jointly guided learning.

Besides, Table 3 summaries the results of our VSG-CNN and 8 leading VD-Augmentation G-ZSL methods. Note that these methods focus more on the generation of synthetic instances of unseen classes, which is time-consuming \cite{long2018pseudo}. Furthermore, the synthetic instances generated are not always reliable due to the visual-semantic gap. In contrast, though our VSG-CNN just exploit the more reliable seen class instances, it still achieves significant performance improvements. As shown in Table 3, our VSG-CNN gains significant advantages in CUB (\textbf{57.0}\% vs\textbf{ 54.3}\%), AWA2 (\textbf{67.0}\% vs \textbf{64.2}\%), and aPY (\textbf{34.0}\% vs \textbf{25.5}\%), though failing on SUN (\textbf{30.9}\% vs \textbf{40.6}\%).

%

\subsubsection{Generalized Open Set Recognition}
Introducing the semantic information from known classes to OSR, this paper first attempt to explore the new generalized open set recognition (G-OSR).

\textbf{Performance Evaluation}.
 As a concept proof that VSG-CNN can effectively utilize the semantic information of seen classes obviously ignored by other existing OSR algorithms, we here only compare VSG-CNN with CPL \cite{yang2018robust} to illustrate this problem. Similar to the $H$ in the evaluation of G-ZSL, we adopt the harmonic mean of known classes' accuracy and unknown class accuracy (correctly rejecting unknown instances). Table 4 shows the results on the four benchmark datasets. Compared with CPL, our VSG-CNN wins on the CUB, SUN, and aPY datasets, while being comparable on AWA2 dataset. To our best knowledge, VSG-CNN is the first method which attempts to effectively use the semantic/attribute information from the known classes in OSR.
\begin{table}[t]
\centering
\caption{Comparison of OSR methods. Best results (\%) are indicated in bold.}
\tabcolsep 4mm
\begin{tabular}{lcccc}
\toprule
\textbf{Methods / Dataset}  &  CUB  & AWA2   & SUN  & aPY  \\
\midrule
  CPL  \cite{yang2018robust} & 70.5   & \textbf{84.5} & 40.3 & 73.6 \\
  VSG-CNN (ours)                 & \textbf{73.3}   & \textbf{84.3} & \textbf{42.3} & \textbf{74.8} \\
\bottomrule
\end{tabular}
\label{tab:booktabs}
\end{table}

\begin{figure*}[!t]
\centering
\subfigure[]{\includegraphics[width=0.32\textwidth]{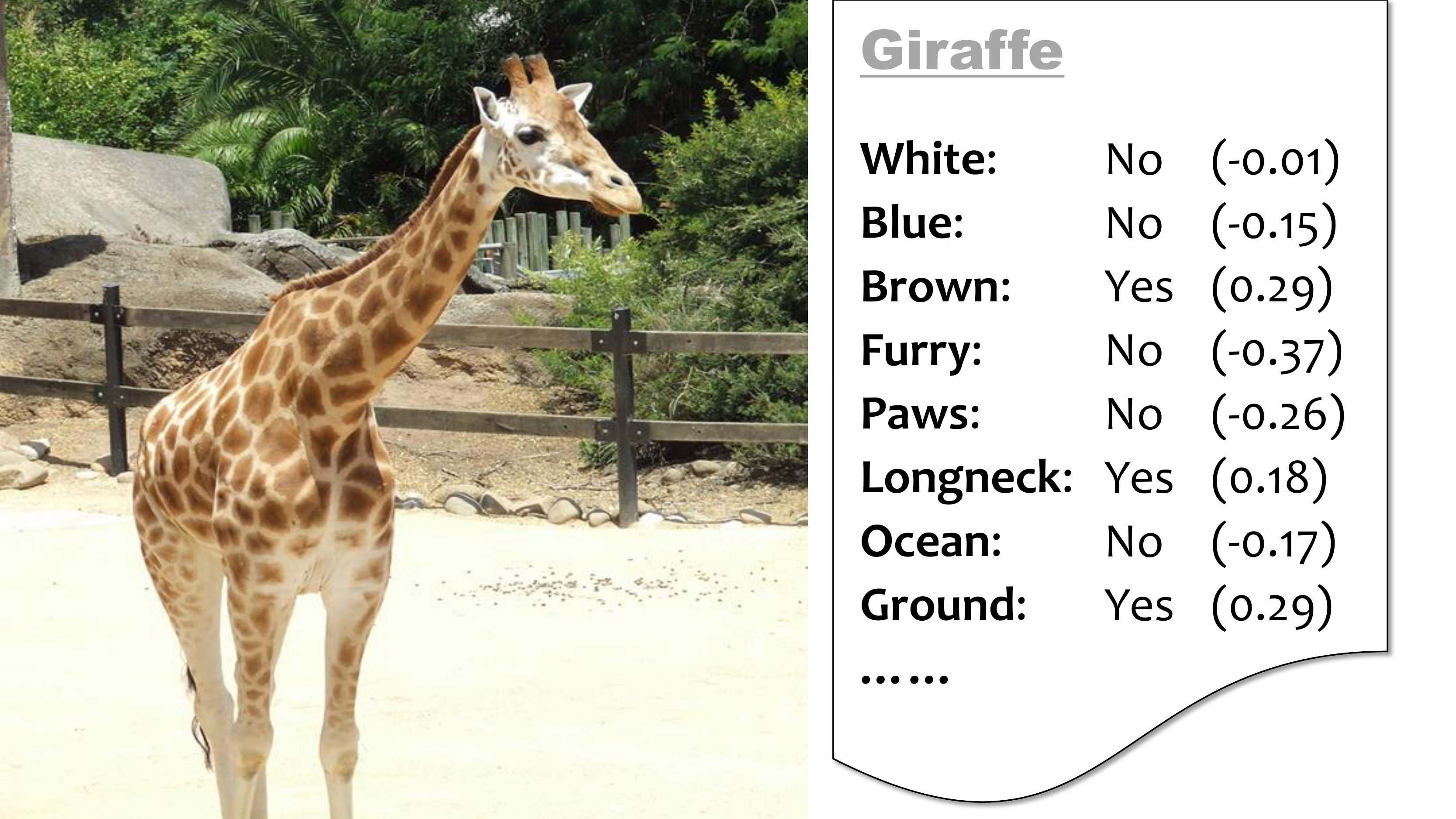}%
\label{fig_first_case}}
\hfil
\subfigure[]{\includegraphics[width=0.32\textwidth]{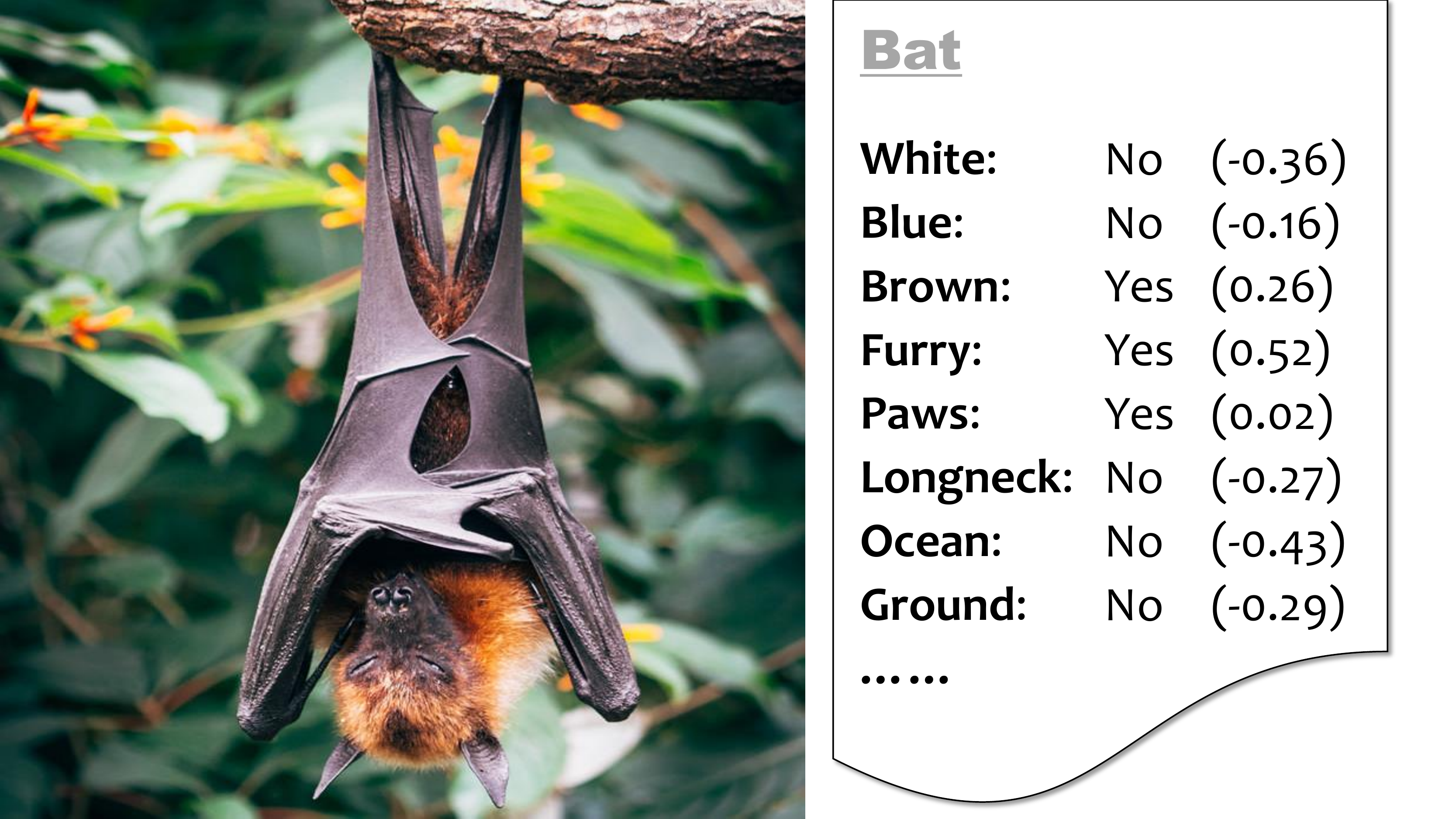}%
\label{fig_second_case}}
\subfigure[]{\includegraphics[width=0.32\textwidth]{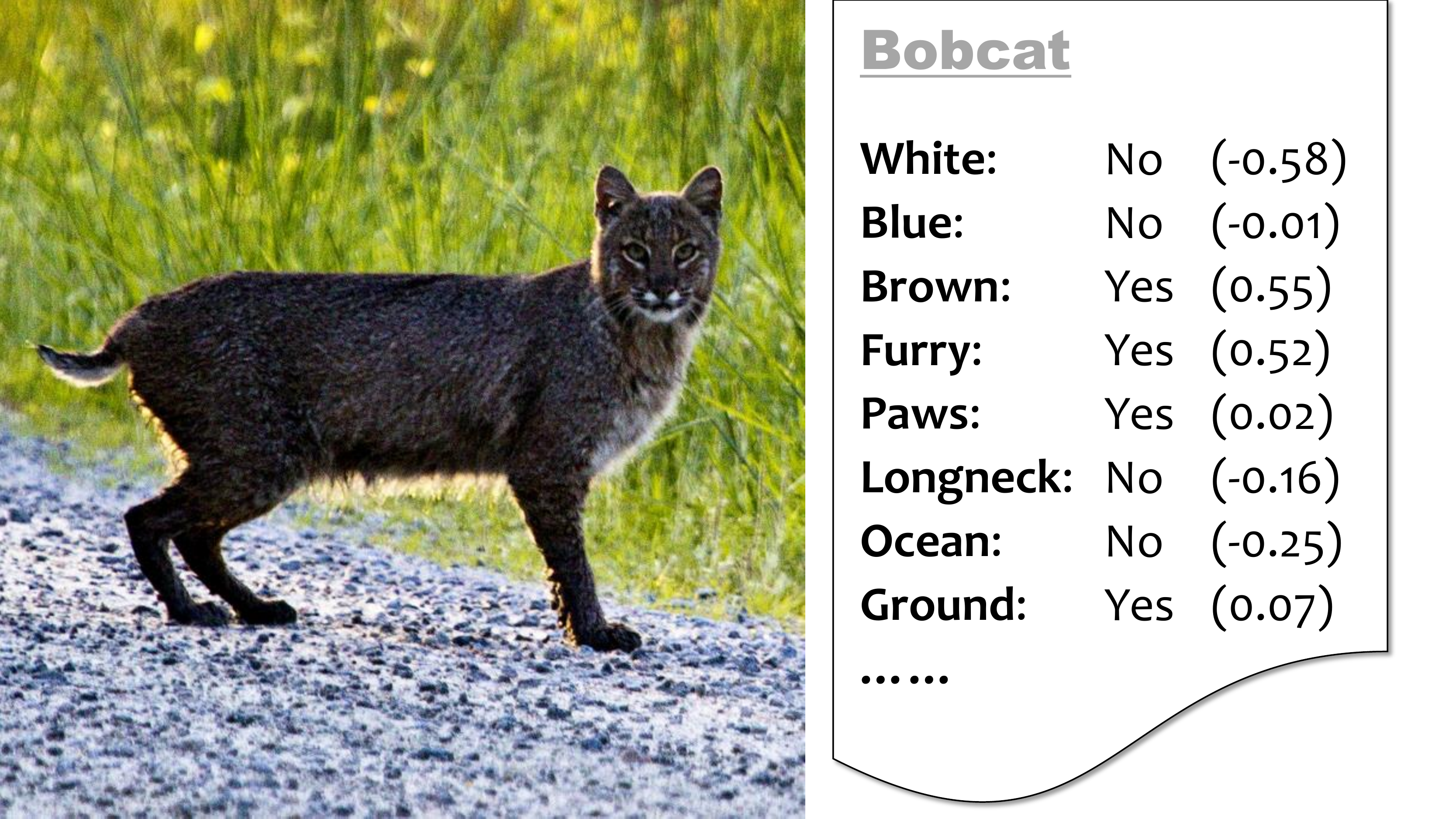}%
\label{fig_first_case}}
\hfil
\subfigure[]{\includegraphics[width=0.32\textwidth]{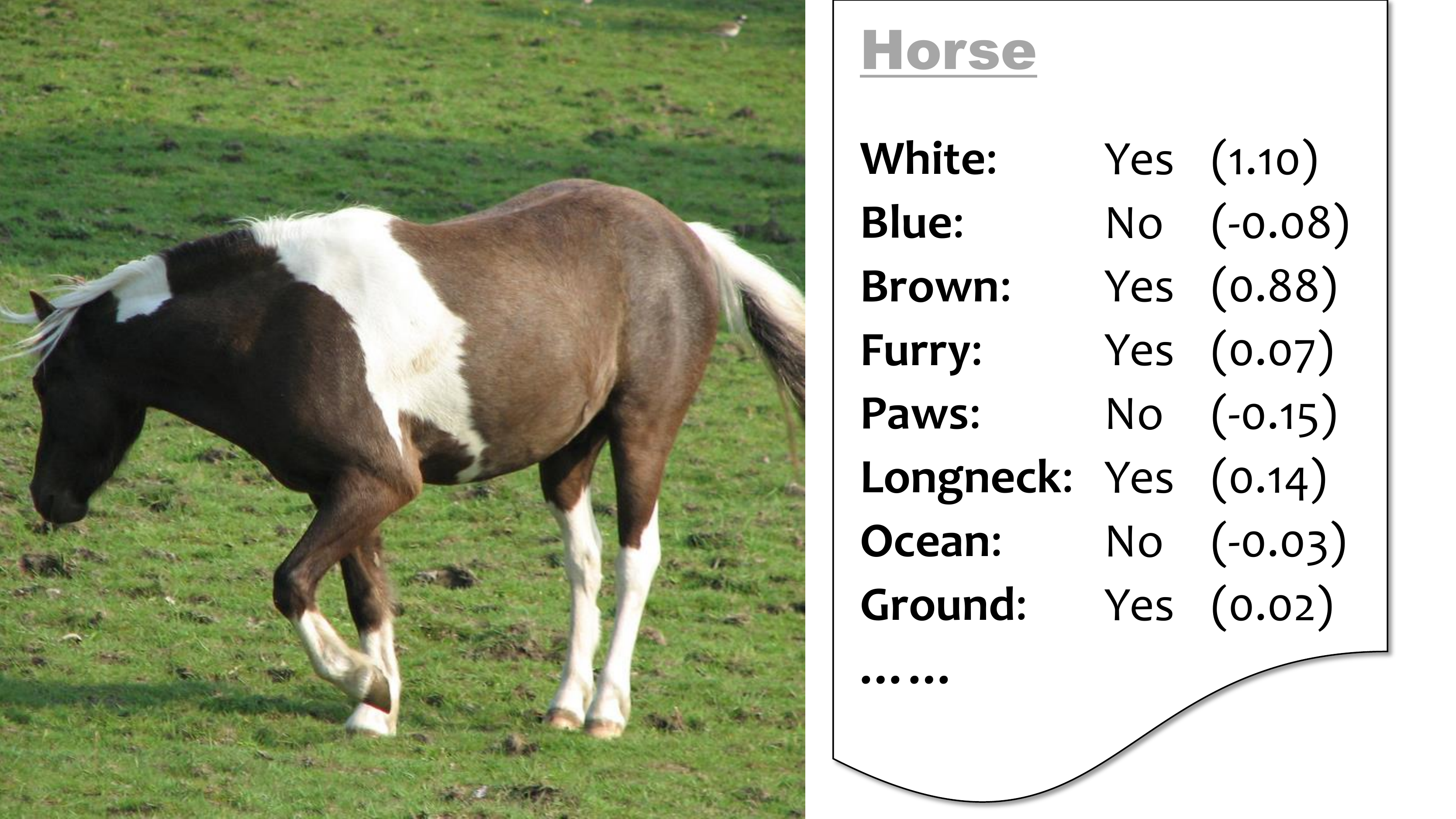}%
\label{fig_second_case}}
\subfigure[]{\includegraphics[width=0.32\textwidth]{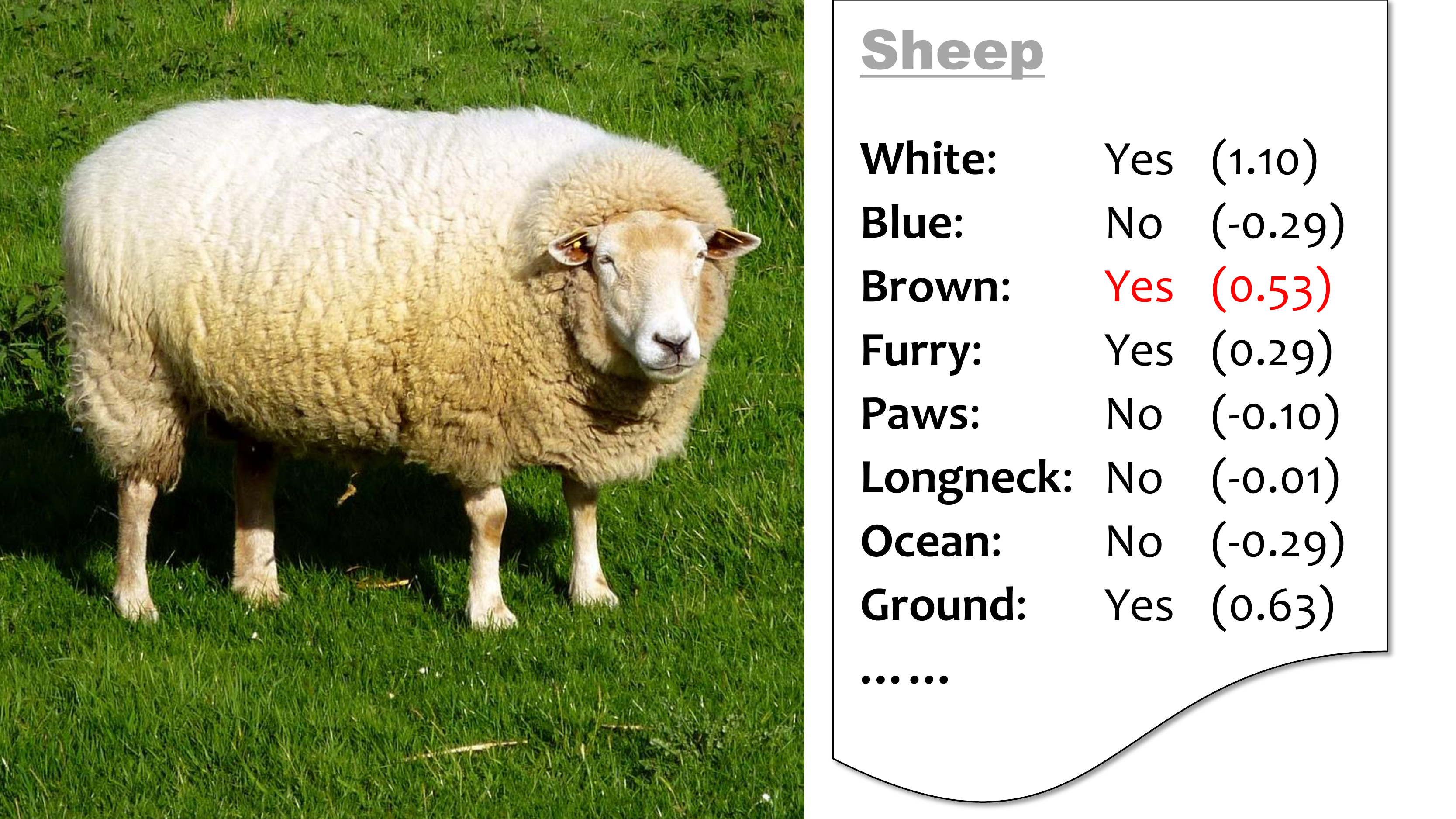}%
\label{fig_first_case}}
\hfil
\subfigure[]{\includegraphics[width=0.32\textwidth]{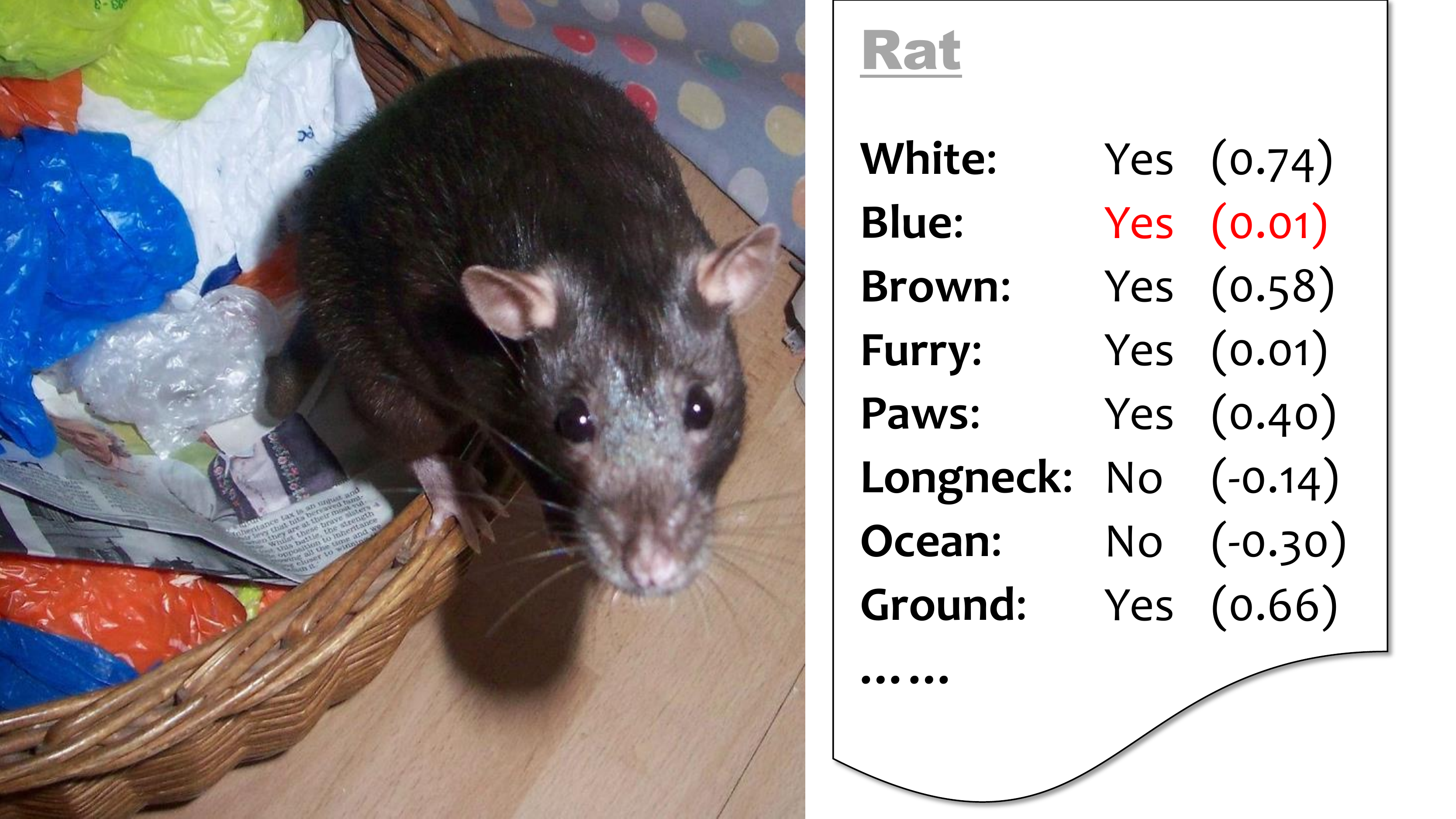}%
\label{fig_second_case}}
\subfigure[]{\includegraphics[width=0.32\textwidth]{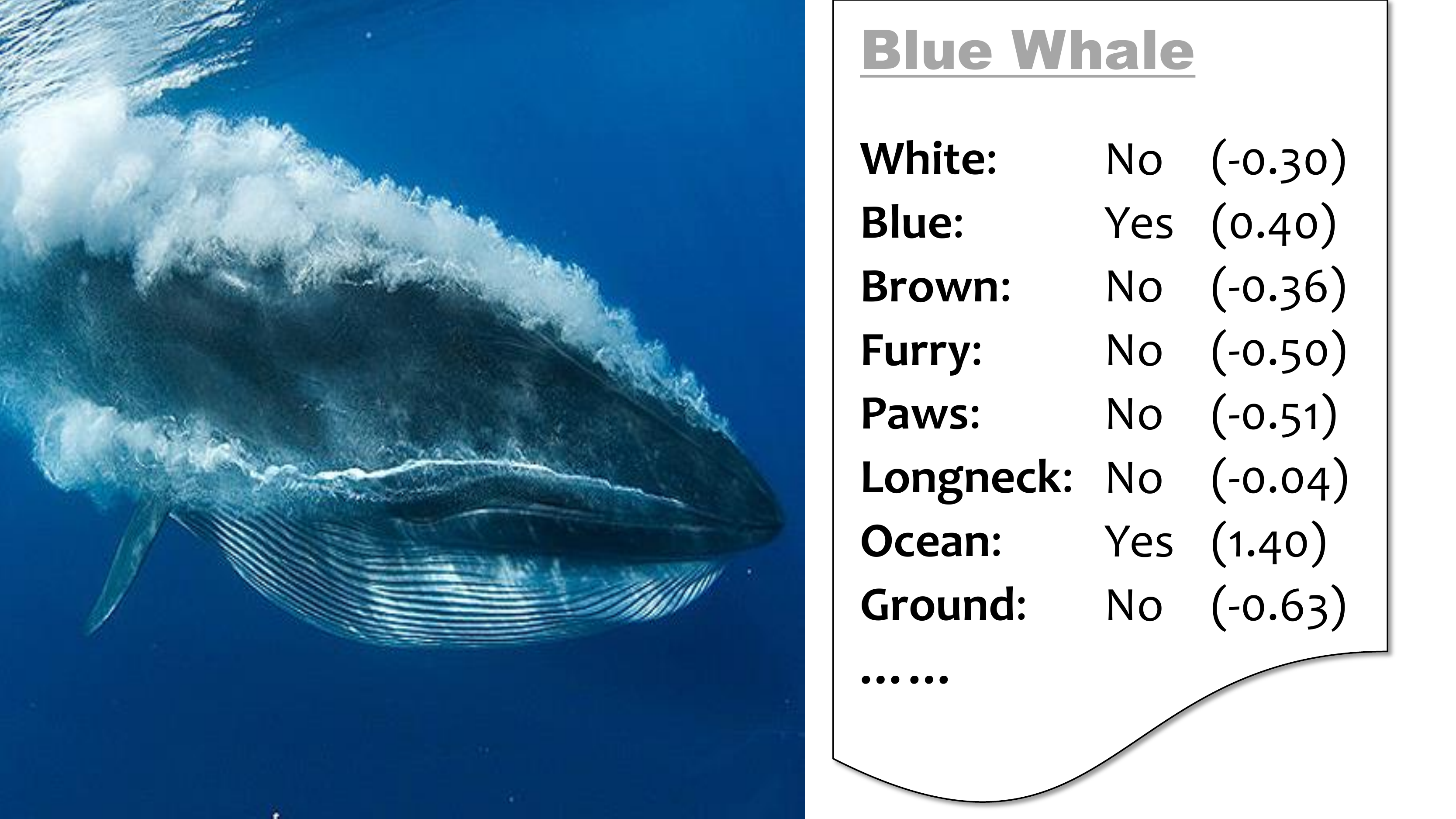}%
\label{fig_first_case}}
\hfil
\subfigure[]{\includegraphics[width=0.32\textwidth]{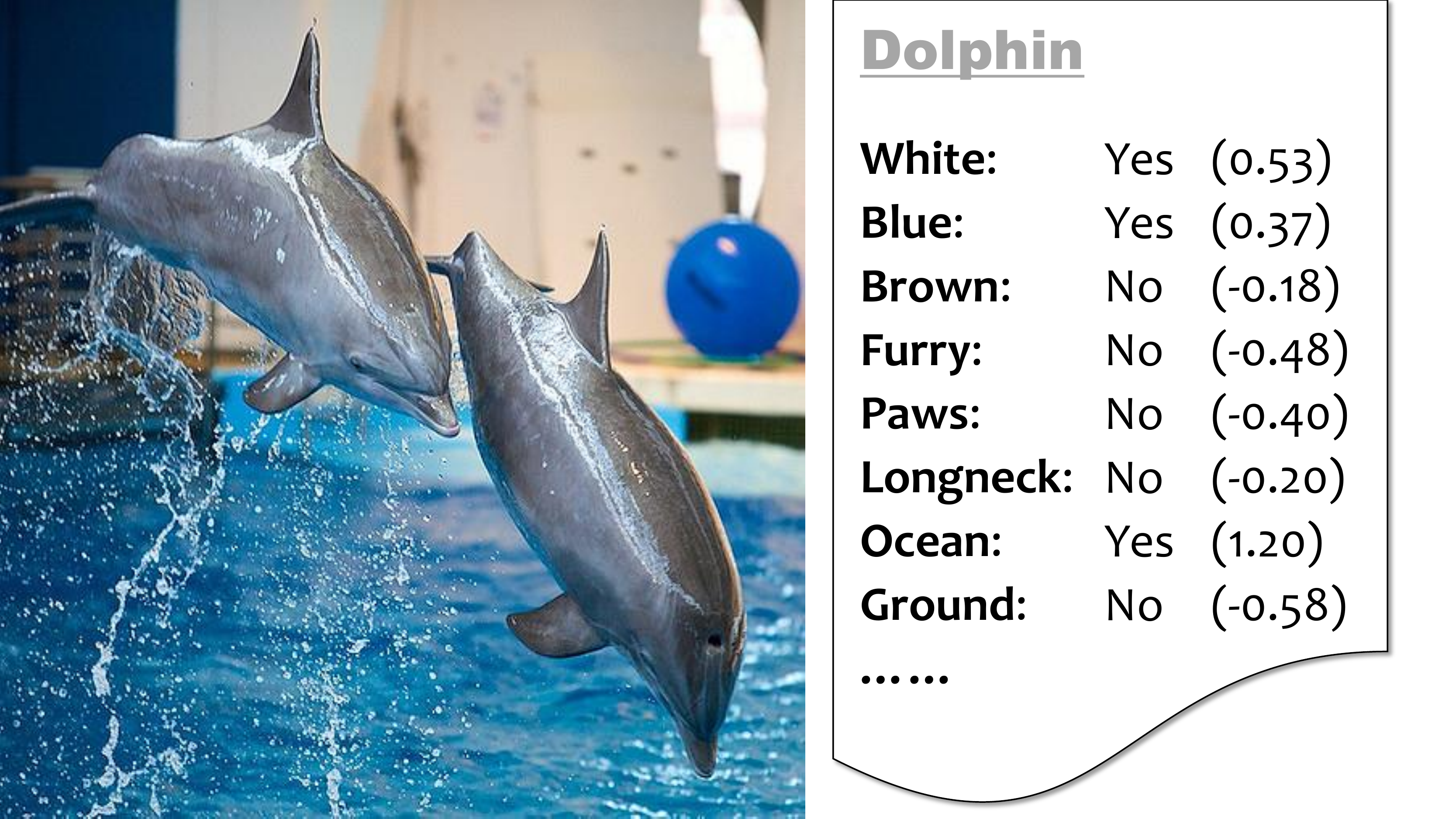}%
\label{fig_second_case}}
\subfigure[]{\includegraphics[width=0.32\textwidth]{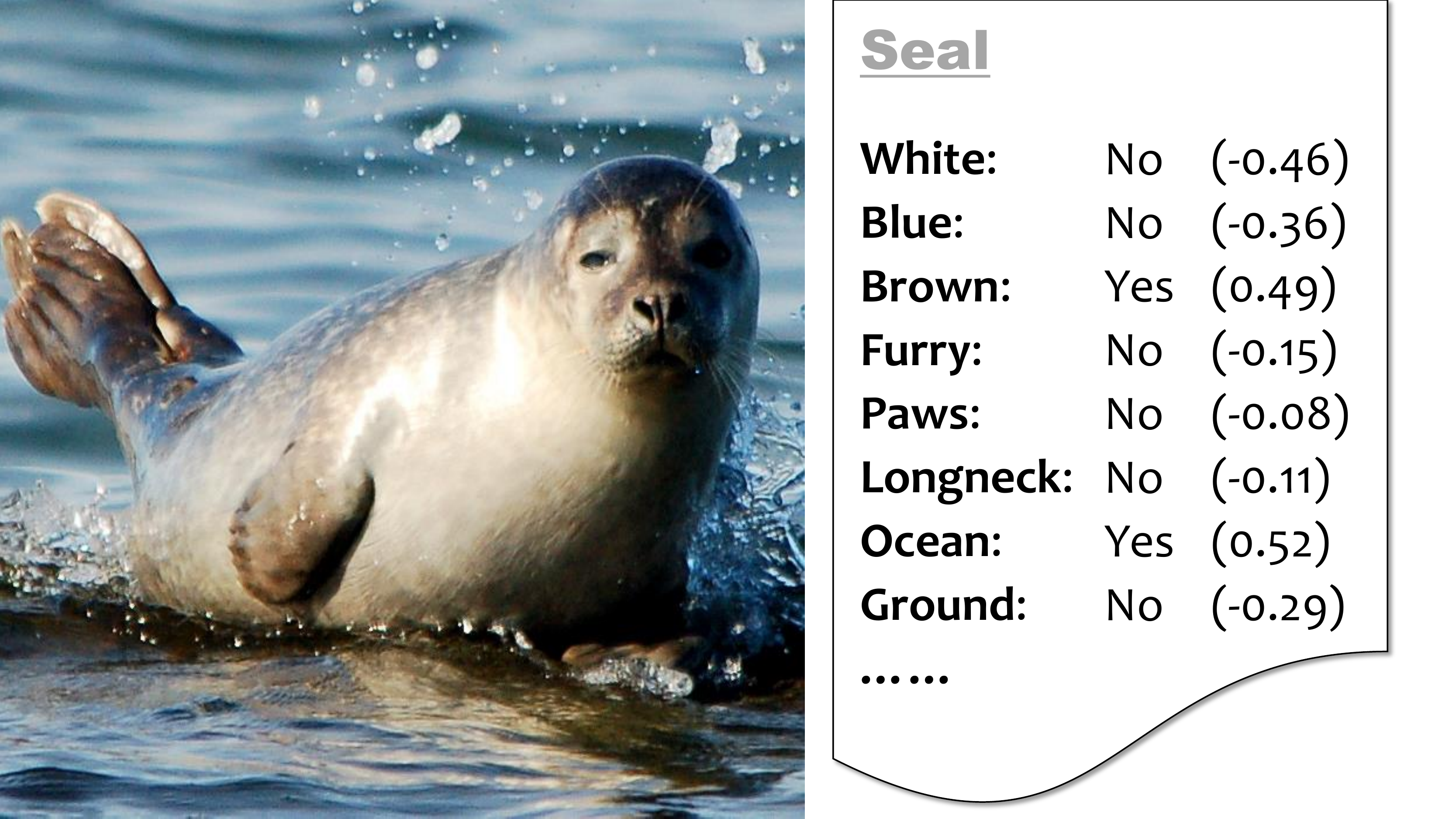}%
\label{fig_second_case}}
\caption{Cognizing unknown class. VSG-CNN utilizes the semantic/attribute information from seen classes to provide a rough semantic/attribute description for the rejected instance. The real values in brackets denote the obtained attribute representation of this rejected instance, where a positive value indicates it has the corresponding attribute, and vice versa. The value marked in red indicates the wrong prediction relative to the ground truth.}
\label{fig_sim}
\end{figure*}

\textbf{Cognizing Unknown Class}. As discussed in Section I, the existing OSR should NOT just rest on a reject decision for unknown class instances but should go forward further. Since a lot of semantic/attribute information (like color \{white, blue, brown, etc.\}, furry \{yes or no\}, ocean\footnote{This attribute indicates the animal lives in ocean  or not.} \{yes or no\}, etc.) is usually shared between the known and unknown classes, we also attempt to exploit this information to provide rough semantic/attribute descriptions for the corresponding rejected instances. Some representative images from the unseen classes in AWA2 are shown in Fig. 5. From these examples, we can see that VSG-CNN can not only reject the unknown class instances but also provide a rough semantic/attribute description for them. Take Fig. 5(a) as an example, based on the semantic prototype vector obtained by VSG-CNN, we can roughly cognize that this new class instance is brown and has spots, longleg, longnech but no stripes, while it lives on ground rather than in ocean. Furthermore, we can also perform coarse-grained category division on the reject instances according to some representative attributes. For example, according to the attributes 'Ocean' and 'Ground', Fig. 5(a,c,d,e,f) can be classified into one category, Fig. 5(g,h,i) the second category, while Fig. 5(b) the third category. Note that the predictions of color attribute in Fig. 5(e,f) are wrong relative to the ground truth. We conjecture it may be the background color in the corresponding images that misleads the classifier.


\section{Conclusion}
Inspired by the process of our humans perceiving the world, this paper decomposes G-ZSL into an OSR and a ZSL tasks so that the seen and unseen classes can be identified separately, effectively solving the CO problem which is ubiquitous in most existing G-ZSL methods. Simultaneously,  without violating OSR's assumptions, we first attempt to explore a new generalized open set recognition (G-OSR) by introducing the semantic information of known classes. Extensive experiments verify the effectiveness of our VSG-CNN. In addition, though the improvement in G-OSR seems incremental in nature, it is worth pointing out that the idea, which utilizes various available information related to known classes rather than just being limited to the visual feature level, undoubtedly provides a new promising direction for expanding the existing OSR research. In fact, a lot of such information, like the knowledge graph \cite{lonij2017open}, the background class data \cite{dhamija2018reducing}, the universum class data\cite{weston2006inference}, etc.,  is often available at hand. We therefore foresee a more generalized setting will be adopted by the future open set recognition.

\section*{Acknowledgments}
The authors would like to thank the support from the
Key Program of NSFC under Grant No. 61732006, NSFC
under Grant No. 61672281, and the Postgraduate Research
\& Practice Innovation Program of Jiangsu Province under
Grant No. KYCX18\_0306.

%
%

%
%
%
%
\bibliographystyle{ieeetr}
\bibliography{mybibfile}



\end{document}